\documentclass[conference]{IEEEtran}
\IEEEoverridecommandlockouts
\usepackage{cite}
\usepackage{amsmath,amssymb,amsfonts}
\usepackage{algorithmic}
\usepackage{graphicx}
\usepackage{textcomp}
\usepackage{xcolor}
\usepackage{booktabs}
\usepackage{multirow}
\usepackage{arydshln}

\def\BibTeX{{\rm B\kern-.05em{\sc i\kern-.025em b}\kern-.08em
    T\kern-.1667em\lower.1ex\hbox{E}\kern-.125emX}}

\IEEEoverridecommandlockouts
\usepackage{bm}
\usepackage{cite}
\usepackage{amsmath,amssymb,amsfonts}
\usepackage{algorithmic}
\usepackage{graphicx}
\usepackage{threeparttable}
\usepackage{textcomp}
\usepackage{algorithmic}
\usepackage{algorithm}
\usepackage{url}
\usepackage{xcolor} 
\usepackage{times}  
\usepackage{mathptmx}

\makeatother

\begin{document}
\title{\LARGE \bf{Anti-Slip AI-Driven Model-Free Control with Global Exponential Stability in Skid-Steering Robots}}
\author{Mehdi Heydari Shahna, Pauli Mustalahti, and Jouni Mattila
\thanks{This work was supported by the Business Finland Partnership Project, `Future All-Electric Rough Terrain Autonomous Mobile Manipulators' under Grant No. 2334/31/2022. (Corresponding author: Mehdi Heydari Shahna.)}
\thanks{The authors are with the Faculty of Engineering and Natural Sciences, Tampere University, Tampere, Finland
        (e-mail: mehdi.heydarishahna@tuni.fi; pauli.mustalahti@tuni.fi; and jouni.mattila@tuni.fi).}
}

\maketitle

\begin{abstract}
Undesired lateral and longitudinal wheel slippage can disrupt a mobile robot's heading angle, traction, and, eventually, desired motion. This issue makes the robotization and accurate modeling of heavy-duty machinery very challenging because the application primarily involves off-road terrains, which are susceptible to uneven motion and severe slippage. As a step toward robotization in skid-steering heavy-duty robot (SSHDR), this paper aims to design an innovative robust model-free control system developed by neural networks to strongly stabilize the robot dynamics in the presence of a broad range of potential wheel slippages. Before the control design, the dynamics of the SSHDR are first investigated by mathematically incorporating slippage effects, assuming that all functional modeling terms of the system are unknown to the control system. Then, a novel tracking control framework to guarantee global exponential stability of the SSHDR is designed as follows: 1) the unknown modeling of wheel dynamics is approximated using radial basis function neural networks (RBFNNs); and 2) a new adaptive law is proposed to compensate for slippage effects and tune the weights of the RBFNNs online during execution. Simulation and experimental results verify the proposed tracking control performance of a $\bm{4,836}$ kg SSHDR operating on slippery terrain.
\end{abstract}


\section{Introduction}
\begin{figure}[h!] 
    \centering
    \scalebox{1.00}{\includegraphics[trim={0cm 0.0cm 0.0cm 0cm},clip,width=\columnwidth]{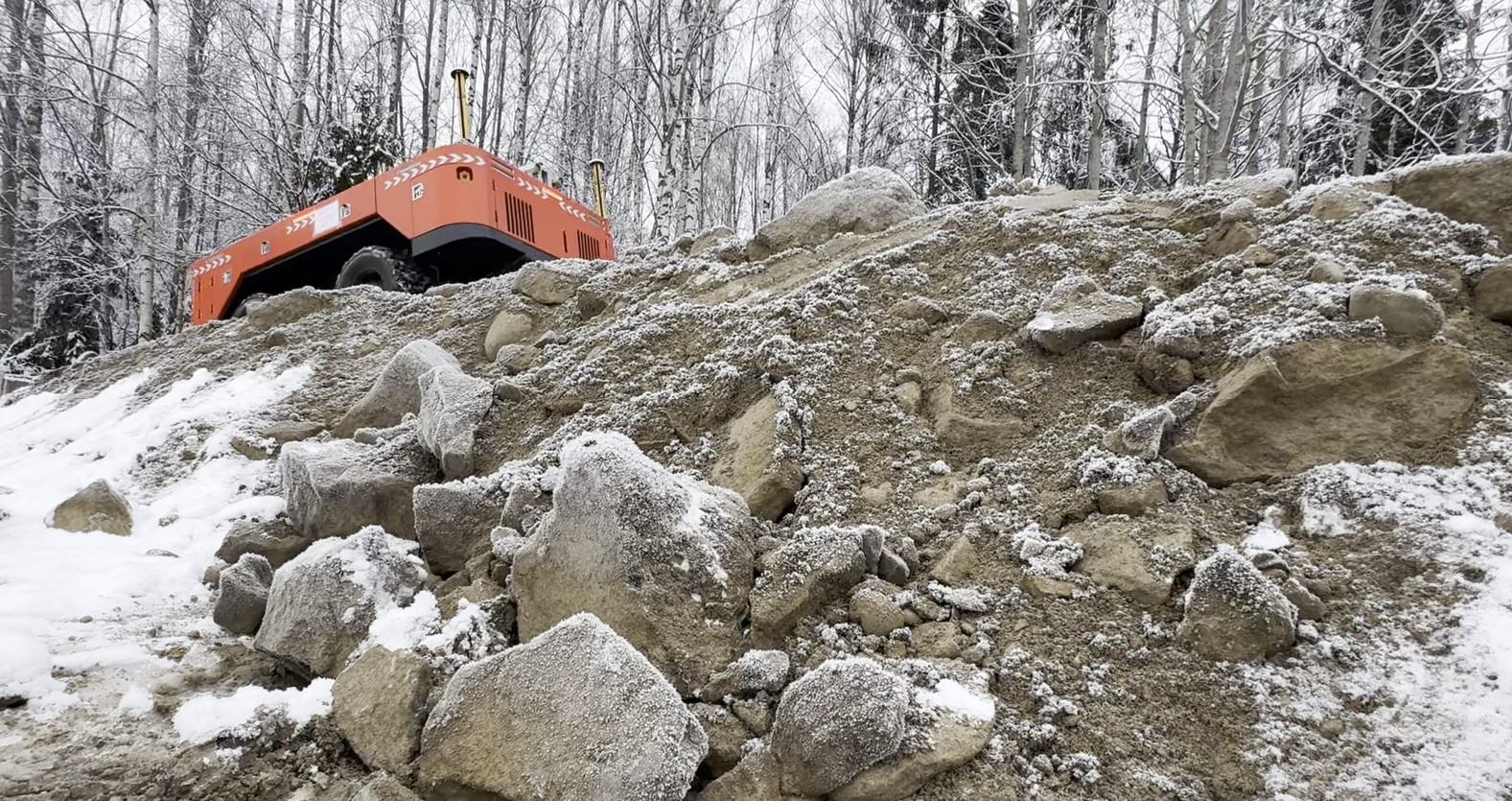}}
    \caption{A four-wheel SSHDR prototype operating on rough terrain.}
    \label{fig4xs5678g}
\end{figure}

The past few years have been marked by significant global changes driven by events such as a worldwide pandemic, supply chain disruptions, resource shortages, and climate change. These factors have accelerated technological shifts, particularly through advancements in robotization and artificial intelligence (AI), which are transforming various business and industrial sectors \cite{machadopath}. One such sector undergoing transformation is the heavy-duty machinery industry, which now requires new insights into autonomy, reliability, and process efficiency for its future development \cite{khan2022overview}.
Robotizing heavy-duty machinery requires advanced solutions that encompass both theoretical developments and application-oriented approaches, along with real-world experimental validations, to enhance automation and safety.
As a result, the development of skid-steering heavy-duty robots (SSHDRs) without a separate steering mechanism has gained increasing demand in industries such as construction, agriculture, and mining. These robot platforms with differential drive systems can perform smallest radius turns by moving one set of wheels forward and the other in reverse, allowing them to pivot within their own length. This capability makes them highly maneuverable and ideal for compact, powerful, and agile operations in confined spaces, but also prone to sliding \cite{aguilera2016general}--\cite{rigotti2020mathcal}. For instance, a four-wheel SSHDR prototype operating on rough terrain is shown in Fig. \ref{fig4xs5678g}.

\begin{figure}[h] 
  \centering
\scalebox{1}
    {\includegraphics[trim={0cm 0.0cm 0.0cm
    0cm},clip,width=\columnwidth]{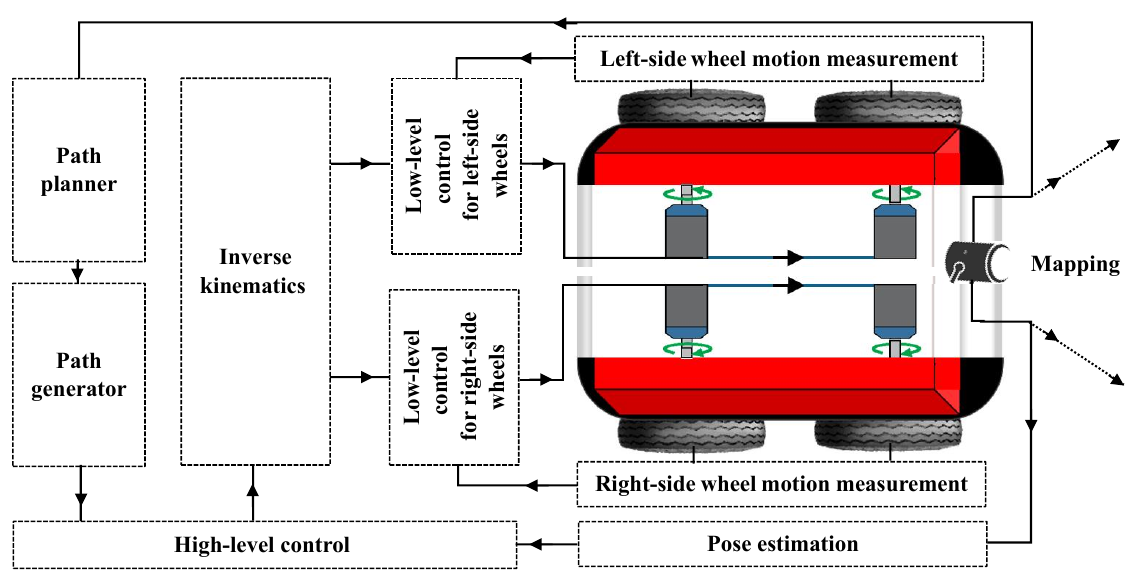}}
  \caption{The whole framework of a four-wheel SSHDR.}
  \label{proto}
\end{figure}

These SSHDRs commonly involve a continuous, real-time engagement of different stages within a framework \cite{liikanen2019path}, as shown in Fig. \ref{proto}. 

\begin{figure}[h] 
  \centering
\scalebox{0.75}
    {\includegraphics[trim={0cm 0.0cm 0.0cm
    0cm},clip,width=\columnwidth]{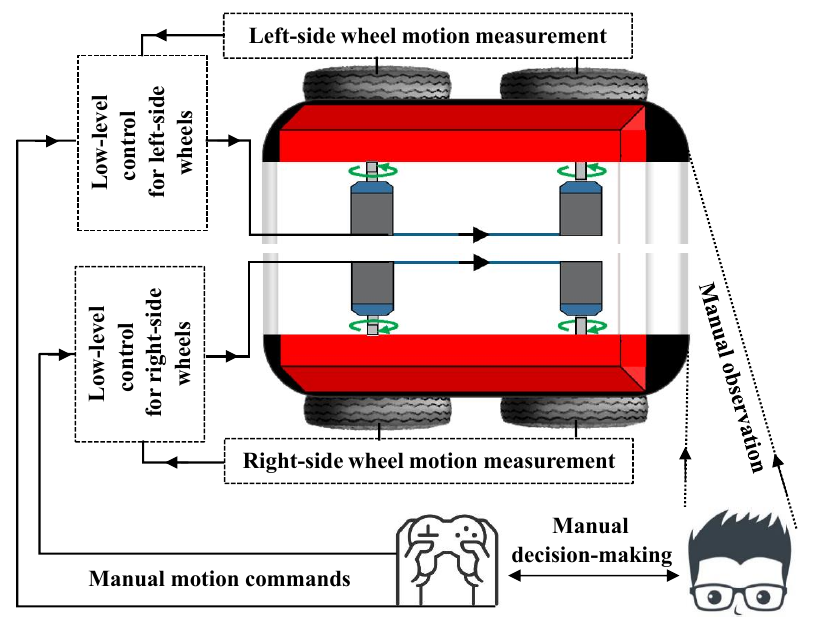}}
  \caption{Human-in-loop SSHDRs for the remote control verification.}
  \label{prosxcdcsfto}
\end{figure}

A representation of the off-road environment is mapped, often aided by sensors such as LiDAR, cameras, inertial measurement units, and GPS for autonomous navigation and localization. This representation is used to estimate the robot’s position and orientation, and the reference motion sector includes a path planner to determine an optimal, collision-free route from start to goal. The path is then refined by a path generator into a smooth, feasible trajectory, considering the robot's kinematics and dynamics. Based on this, the robot's required linear and angular velocities, from the immediate pose to the goal, are computed by kinematic control. The required robot motion is then transformed into individual left and right reference wheel motions using inverse kinematics.
Integrating these complex and broad sectors, each requiring advanced methods to operate, with the expectation that they will seamlessly contribute to the final objective, is unrealistic in rough terrains. They must be carefully designed and separately tested to meet various standards before integration. Specifically, one common approach to evaluate the efficacy of dynamic controllers within the overall SSHDR framework in industries is through human-assisted observation and manual decision-making in the closed-loop control system, as shown in Fig. \ref{prosxcdcsfto}.

\begin{figure}[h] 
  \centering
\scalebox{1}
    {\includegraphics[trim={0cm 0.0cm 0.0cm
    0cm},clip,width=\columnwidth]{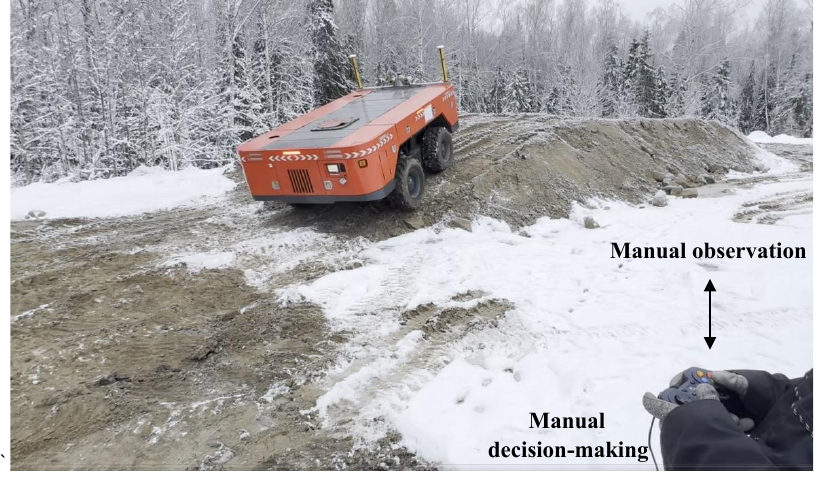}}
  \caption{Verification of the remote control system in practice; the video is available at: \texttt{https://youtu.be/AQycbEVt\_M8}.}
  \label{fvdproto}
\end{figure}

This allows a heavy robot to operate in challenging terrains with a high likelihood of slippage while enabling the human operator to assess obstacles along the path, the robot's distance to the destination, and approximate motion commands in real-time. One instance of verifying execution of the control system in practice is shown in Fig. \ref{fvdproto}.

Dynamic controllers are the lowest executive system, responsible for robustly adjusting the multi-stage mechanical actuator mechanism to generate sufficient motion and ensuring stability and goal achievement. 
The actuation systems integrated into SSHDRs typically consist of multi-component mechanisms designed to transform energy from the power source into mechanical motion. This process involves energy excitation, high-torque motor operation, high-ratio gearing, and, ultimately, the generation of wheel movement. The inherent complexity and nonlinearities of this multi-stage actuation system pose significant challenges in accurately modeling the SSHDR's dynamics \cite{shahna2024model}. Consequently, designing a model-based controller is significantly complicated by substantial modeling uncertainties and external disturbances. Hence, traditional model-free controllers remain widely used in the heavy-duty machinery industry due to their simplicity and lower computational demands, despite advancements in intelligent control methodologies. However, the performance of the common model-free controllers in higher-than-second-order systems is often limited, leading to oscillations, poor performance, or instability \cite{zhao2017pid}. 
In addition, the safety and stability of SSHDRs on uneven terrain are significantly affected by wheel slippage, which arises from factors such as terrain irregularities, imbalanced load distribution, suboptimal wheel design, dynamic forces during movement, and limitations in the robot’s control mechanisms that may hinder timely adjustments \cite{wang2008modeling}.
Notably, despite the critical impact of wheel slippage, research on this issue across all types of heavy-duty robots has received limited attention over the past two decades \cite{teji2023survey}--\cite{shahna2024robust}. Various terrains with differing slip ratios must be thoroughly analyzed to ensure the stability of the robot’s autonomous controller before integrating it with mapping systems and obstacle-free trajectory generators. Hence, \cite{liao2017performance} integrated a coordinated adaptive robust control scheme with a torque allocation technique to solve the chattering phenomenon and wheel slip compensation. However, it assumed the model of the studied skid-steering robot is accurately known. To improve the dynamic controller performance and tracking accuracy,  \cite{xiong2022path} developed a traditional controller, the cascaded proportional–integral–derivative (PID) method, while compensating for the sideslip angles. However, most studies have been built on the assumption that all four wheels of the studied robot can be controlled independently, which makes it easier to address the slipping issue and compensate for it.

This paper addresses the problem of robust stabilization for a four-wheel SSHDR system, in which each two-wheel side can be controlled with a single actuator. The system is characterized by a highly nonlinear multi-component actuation model and operates on severely slippery terrains. 
In contrast to \cite{liao2017performance} and \cite{xiong2022path}, the proposed control framework is entirely independent of the SSHDR model and interactive multi-stage actuation mechanisms. In addition, unlike \cite{liikanen2019path}, the robustness of the proposed control is guaranteed, adaptively ensuring operation on rough terrains with varying slip ratios.
Within the proposed control framework, the functional modeling of the wheel dynamics is approximated using radial basis function neural networks (RBFNNs), where, in contrast to \cite{yang2011adaptive} and \cite{tong2011adaptive}, the NN weights, approximation errors, and their norms are not required to be known. The proposed novel adaptive algorithm updates these values and is also responsible for compensating for disturbances caused by wheel slippage. By incorporating the online-updated values of uncertain functions and parameters into the control framework, for the first time in SSHDRs, model-free tracking of the commanded motion with global exponential stability is ensured. Global exponential stability, a strong form of stability in applications, implying uniform, asymptotic stability with exponential convergence rate, holds for all initial conditions. 
This work lays the foundation for advancing reliable autonomous navigation in heavy-duty machinery undergoing robotization, as shown in Fig. \ref{proto}.

The rest of this paper begins with Section II to provide a mathematical evaluation of the effects of rough terrains with varying slip ratios on the differential drive system dynamics of the SSHDR. In Section III, the governing dynamics of the slip-affected SSHDR are utilized to design a robust model-free control developed by a neural networks (NN-RMFC) system independently for each side. Section IV provides rigorous stability proof for the proposed controller. In Section V, a guide for tuning control design parameters is provided. Finally, in Section VI, the control method is applied both theoretically and experimentally to the SSHDR operating in high-slip-ratio terrain to demonstrate the effectiveness of trajectory tracking.

\section{Differential Drive System Dynamics}
Let us define the motion dynamics of both wheeled sides, right (R) and left (L) as follows: 
\begin{equation}
\small
\begin{aligned}
\label{e1}\dot{V}_i =& g_{i} U_i+d_i\left(\bm{V}\right)+\Delta_{i}(t)
\end{aligned}
\end{equation}
where $i \in\{R, L\}$, $V_i$ is the linear velocity of each wheeled side, and $\bm{V}=[V_R, V_L]^\top \in \mathbb{R}^2 \rightarrow \mathbb{R}$. $U_i: \mathbb{R} \rightarrow \mathbb{R}$ is the control input, $d_i(\cdot): \mathbb{R} \times \mathbb{R}^2 \rightarrow \mathbb{R}$ represents the uncertain multi-stage actuation model of the robot, $g_i \in \mathbb{R}^+$ is the unknown control coefficient due to the inertia of the robot, and $\Delta_i(t): \mathbb{R} \rightarrow \mathbb{R}$ represents external disturbances. The disturbance structure under slippage can be changed to $\Delta_i(t)=-{\mu_s}_i(s) F_{id}$, where $F_{id}\in \mathbb{R} \rightarrow \mathbb{R}$ represents external disturbance under no-slip conditions, and ${\mu_s}_i(s)=1+s_i$ denotes the functional slippage effect derived from \cite{andreev2020trajectory}. The slip ratio of the wheel ($s_i \in \mathbb{R} \times \mathbb{R} \rightarrow \mathbb{R}$) can be defined as \cite{jiang2022adaptive}:
\begin{equation}
\small
\begin{aligned}
\label{e2}
s_i=\frac{({\omega_\omega}_i- \omega_i)}{\max({\omega_\omega}_i, \omega_i)}, \hspace{0.2cm} |s_i|<1
\end{aligned}
\end{equation}
where ${\omega_{\omega}}_i \in \mathbb{R} \rightarrow \mathbb{R}$ and $\omega_i \in \mathbb{R} \rightarrow \mathbb{R}$ are the wheel’s theoretical and actual angular velocity, respectively. Note $V_i = r_i \omega_i $. As we assume that the SSHDR dynamic model is unknown, from Eq. \eqref{e1}, the only information available is the velocity of the wheeled side of the robot $V_i$ obtained from motion sensors. The aim is to design control input $U_i$ for the robot to track the manual motion commands.
\section{Rbfnn-Based NN-RMFC}
\begin{figure*} [t]
    \centering
    \includegraphics[width=0.8\textwidth, height=9.5cm]{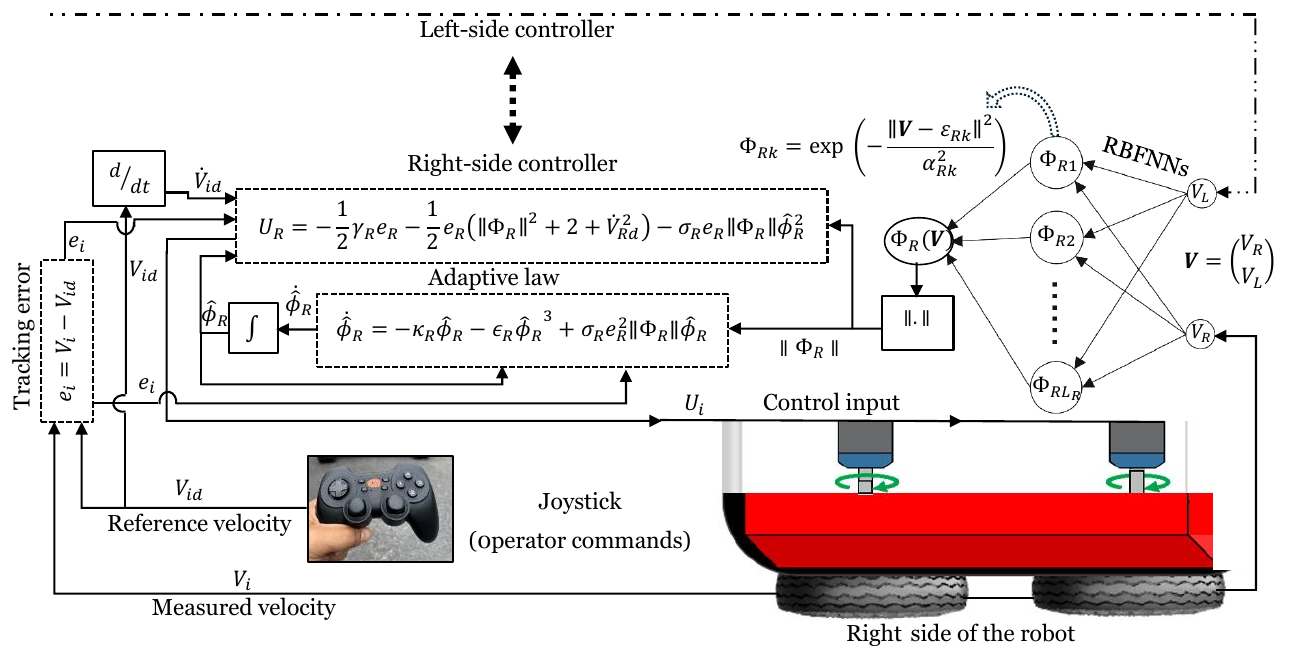}
    \caption{The right side of the control-implemented SSHDR as a representative sample. The left side follows the same structure.}
    \label{fi121}
\end{figure*}

The tracking error is defined as $e_{i} =  V_i - V_{id}$ where $V_{id}$ is the commanded linear velocity of each side, generated by human input. From \eqref{e2}, we have
\begin{equation}
\small
\begin{aligned}
\label{e3}
\dot{e}_{i} =  g_{i} U_i+d_i\left(\bm{V}\right)+\Delta_{i}(t) - \dot{V}_{id}(t)
\end{aligned}
\end{equation}
Let us form the unknown modeling system term $d_i$ utilizing RBFNNs as
\begin{equation}
\small
\begin{aligned}
\label{e4}
d_i\left(\bm{V}\right)=\left(a_i\right)^{\top} \Phi_i\left(\bm{V}\right)+\delta_i\left(\bm{V}\right)
\end{aligned}
\end{equation}
where $\bm{V} \in \mathbb{R}^2 \rightarrow \mathbb{R}$ is the input of RBFNNs. $a_i \in  R^{L_i}$ is the weight vector of the output layer with the number of hidden-layer neurons $L_i$, and $\delta_i(\bm{V})  \in \mathbb{R}^2 \rightarrow \mathbb{R}$ is the approximation error. We assume that both $a_i$ and $\delta_i$ as well as their upper bounds are unknown.
$\Phi(.) \in \mathbb{R}^2 \rightarrow R^{L_i}$ is the NN basis function defined as:
\begin{equation}
\small
\begin{aligned}
\label{e5}
\Phi_i\left(\bm{V}\right)=\left[\begin{array}{lll}
\Phi_{i 1}\left(\bm{V}\right) & \Phi_{i 2}\left(\bm{V}\right) \cdots \Phi_{i L_i}\left(\bm{V}\right)
\end{array}\right]^{\top}
\end{aligned}
\end{equation}
$\Phi_{i k}\left(\bm{V}\right)$ can be chosen as the commonly used Guassian functions, such as the following:
\begin{equation}
\small
\begin{aligned}
\label{e6}
\Phi_{i k}\left(\bm{V}\right)=\exp \left(-\frac{\left\|\bm{V}-{\varepsilon}_{i k}\right\|^2}{\alpha_{i k}^2}\right), \quad k=1,2, \ldots, L_i
\end{aligned}
\end{equation}
where $\varepsilon_{i k} \in \mathbb{R}^2$ is any given two-dimensional constant vector representing the center of the receptive field, and $\alpha_{i k} \in \mathbb{R} $ is any given constant representing the width of the Gaussian functions. The notation $\|.\|$ represents the Euclidean norm. Let us propose a novel adaptive law for tuning \eqref{e4} and compensating $\Delta_i$ in \eqref{e3}, as follows:
\begin{equation}
\small
\begin{aligned}
\label{e7}
\dot{\hat{\phi}}_i=&-\kappa_i \hat{\phi}_i-\epsilon_i\hat{\phi}^3_i+\sigma_i {e^2_i} \|\Phi_i\| \hat{\phi}_i
\end{aligned}
\end{equation}
where $\kappa_{i}$, $\epsilon_{i}$, and $\sigma_{i} \in \mathbb{R}$ are positive constants. Now, we propose a novel model-free RFBNN-based adaptive control, as follows:
\begin{equation}
\small
\begin{aligned}
\label{e8}
U_{i}=&-\frac{1}{2} \gamma_i {e}_i-\frac{1}{2}{{e}_i}(\|\Phi_i\|^2+2+\dot{V}^2_{id})-\sigma_i {e_i} \|\Phi_i\| \hat{\phi}^2_i
\end{aligned}
\end{equation}
where $\gamma_i$ is a positive constant. The proposed control framework is illustrated in Fig. \ref{fi121}. This figure presents only the right side of the control-implemented SSHDR as a representative sample, as the left side follows the same structure.

\textit{Assumption} 1: Assume $g_i$ in \eqref{e1} and \eqref{e3} is an unknown positive, meaning that the direction of the control signal and the acceleration of the robot are the same. 

\textit{Assumption} 2: The approximation error of the RBFNNs in Eq. \eqref{e4} and slippage-caused disturbance are bounded and satisfy $\|\delta_i(\bm{V})\| \leq \bar{\delta}_i<\infty$ and $\|\Delta_i(t)\| \leq \bar{\Delta}_i<\infty$, where $\bar{\delta}_i$ and $\bar{\Delta}_i$ are unknown and positive.

\textit{Definition} 1 \cite{murray1993robotic}. Consider a nonlinear system as follows:
\begin{equation}
\small
\begin{aligned}
\label{45275}
\dot{x}=f(x, t) \quad x\left(t_0\right)=x_0 \quad x \in \mathbb{R}^n 
\end{aligned}
\end{equation}
where $f(x, t)$ is piecewise continuous in $t$, uniformly in $t$, and Lipschitz continuous with respect to $x$. For constants $m, \alpha>0$ and $\epsilon>0$, the equilibrium point $x^*$ is a globally exponentially stable equilibrium point if
\begin{equation}
\small
\begin{aligned}
\label{4527adsa5}
\|x(t)\| \leq m e^{-\alpha\left(t-t_0\right)}\left\|x\left(t_0\right)\right\|
\end{aligned}
\end{equation}
for all $x_0 \in \mathbb{R}^n$. $\alpha$ is called the exponential rate of convergence.

\textit{Theorem} 1:
Employing the proposed controller provided in Eq. \eqref{e8} which includes RBFNNs Eq. \eqref{e5} and adaptive laws Eq. \eqref{e7}, for the SSHDR under unknown but bounded systematic modeling and disturbances, ensures the vector of tracking errors $\bm{e}=[e_R, e_L]^\top$ not only converges to zero but does so at an exponential rate, regardless of the initial condition.

{Proof.} See Section IV.

\section{Stability Proof}
Let us define a quadratic function as
\begin{equation}
\small
\begin{aligned}
\label{e9}
\bar{V}_i=&\frac{1}{2{g}_i} e^2_i+\frac{1}{2} \hat{\phi}_i^2
\end{aligned}
\end{equation}
After differentiating Eq. \eqref{e9} and employing Eq. \eqref{e7}, we have
\begin{equation}
\small
\begin{aligned}
\label{e10}
\dot{\bar{V}}_i=& e_i \frac{g_{i}}{{g}_i} U_i+\frac{1}{{g}_i} e_i d_i\left(\bm{x}\right)+\frac{1}{{g}_i} e_i\Delta_{i}(t)\\
&- \frac{1}{{g}_i} e_i \dot{V}_{id}-\kappa_i \hat{\phi}^2_i-\epsilon_i\hat{\phi}^4_i+\sigma_i {e^2_i} \|\Phi_i\| \hat{\phi}^2_i
\end{aligned}
\end{equation}
By inserting Eq. \eqref{e8} and Eq. \eqref{e4} into Eq. \eqref{e10}, we obtain
\begin{equation}
\small
\begin{aligned}
\label{e11}
\dot{\bar{V}}_i=& -\frac{1}{2} \gamma_i {e}^2_i-\frac{1}{2}{{e}^2_i}(\|\Phi_i\|^2+2+\dot{V}^2_{id})\\
&-\sigma_i {e^2_i} \|\Phi_i\| \hat{\phi}^2_i
+ \frac{1}{{g}_i} {e}_i\left(a_i\right)^{\top} \Phi_i+\frac{1}{{g}_i} {e}_i \delta_i \\
&+ \frac{\Delta_i}{{g}_i}{e}_i- \frac{1}{{g}_i} {e}_i \dot{V}_{id}-{\kappa_i}\hat{\phi}^2_i-\epsilon_i \hat{\phi}^4_i\\
&+{\sigma_i} {e_i}^2 \|\Phi_i\| \hat{\phi}^2_i
\end{aligned}
\end{equation}
Considering the simple mathematical relationship  $\frac{1}{2}a^2 + \frac{1}{2}b^2 \geq ab $, disregarding the value of $a$ and $b$, we have
\begin{equation}
\small
\begin{aligned}
\label{e12}
\dot{\bar{V}}_i \leq & -\frac{1}{2} \gamma_i {e}^2_i-\frac{1}{2}{{e}^2_i}(\|\Phi_i\|^2+2+\dot{V}^2_{id})\\
&-\sigma_i {e^2_i} \|\Phi_i\| \hat{\phi}^2_i+ \frac{1}{2{g}^2_i} \|a_i^{\top}\|^2 +\frac{1}{2} {e}^2_i \|\Phi_i\|^2 \\
&+\frac{1}{2{g}^2_i} \delta^2_i+ \frac{1}{2} {e}^2_i +\frac{\Delta^2_i}{2{g}^2_i}+\frac{1}{2}{e}^2_i+ \frac{1}{2{g}^2_i} \\
&+ \frac{1}{2} {e}^2_i \dot{V}^2_{id}-\kappa_i \hat{\phi}^2_i-\epsilon_i \hat{\phi}^4_i+{\sigma_i}{e_i}^2 \|\Phi_i\| \hat{\phi}^2_i
\end{aligned}
\end{equation}
After simplifying Eq. \eqref{e12}
\begin{equation}
\small
\begin{aligned}
\label{e13}
\dot{\bar{V}}_i \leq & -\frac{1}{2}  \gamma_i {e}^2_i-\frac{1}{2}{{e}^2_i}(\|\Phi_i\|^2+2+\dot{V}^2_{id})\\
& +\frac{1}{2} {e}^2_i (\|\Phi_i\|^2+ 1 +1+ \dot{V}^2_{id}) \\
&+ \frac{1}{2{g}^2_i}(1+\Delta^2_i+\delta^2_i+\|a_i^{\top}\|^2) -{\kappa_i} \hat{\phi}^2_i-\epsilon_i \hat{\phi}^4_i
\end{aligned}
\end{equation}
Thus, we easily define $\hat{\phi}_i$ based on \textit{Assumption} 2 as:
\begin{equation}
\small
\begin{aligned}
\label{e14}
\dot{\bar{V}}_i \leq & -\frac{1}{2}  \gamma_i {e}^2_i+ \frac{1}{2{g}^2_i}(1+\bar{\Delta}^2_i+\bar{\delta}^2_i+\|a_i^{\top}\|^2) -{\kappa_i} \hat{\phi}^2_i\\
&-\epsilon_i \hat{\phi}^4_i
\end{aligned}
\end{equation}
Thus, we easily obtain the adaptive parameter supposed to be estimated in \eqref{e7}, as follows:
\begin{equation}
\small
\begin{aligned}
\label{e15}
\hat{\phi}_i=\pm\sqrt[4]{(2 g_i \epsilon_i)^{-1} (1+\|a_i^{\top}\|^2+\bar{\delta}^2_i+\bar{\Delta}^2_i)}
\end{aligned}
\end{equation}
Thus
\begin{equation}
\small
\begin{aligned}
\label{e16}
\dot{\bar{V}}_i\leq&  -\frac{1}{2}  \gamma_i {e}^2_i -{\kappa_i} \hat{\phi}^2_i
\end{aligned}
\end{equation}
From Eq. \eqref{e9}, we have
\begin{equation}
\small
\begin{aligned}
\label{e17}
\dot{\bar{V}}_i\leq&  -\rho_i V_i
\end{aligned}
\end{equation}
where $\rho_i = \min \hspace{0.1cm} [\hspace{0.1cm} {g}_i \gamma_i,\hspace{0.2cm}2{\kappa_i}\hspace{0.1cm}]$. Considering both sides ($i: R, L$), we can define an entirely quadratic function, as follows:
\begin{equation}
\small
\begin{aligned}
\label{e18}
{\bar{V}}=& \bar{V}_R + \bar{V}_L=\\
=&\frac{1}{2{g}_R} e^2_R+\frac{1}{2} \hat{\phi}_R^2+\frac{1}{2{g}_L} e^2_L+\frac{1}{2} \hat{\phi}_L^2
\end{aligned}
\end{equation}
Defining $\bm{e}=\begin{bmatrix}
{e_R}& e_L
\end{bmatrix}^{\top}$, $\bm{\phi}=\begin{bmatrix}
{\phi_R}& \phi_L
\end{bmatrix}^{\top}$, and $\bm{g}=\begin{bmatrix}
\frac{1}{{g}_R} & 0 \\
0 & \frac{1}{{g}_L}
\end{bmatrix}$, we have
\begin{equation}
\small
\begin{aligned}
\label{e19}
{\bar{V}}\leq& \frac{1}{2}\bm{e}^\top \bm{g} \hspace{0.1cm} \bm{e} + \frac{1}{2}\bm{\phi}^{\top} \hspace{0.1cm} \bm{\phi}
\end{aligned}
\end{equation}
From Eq. \eqref{e16}, and differentiating Eq. \eqref{e19}, we obtain
\begin{equation}
\small
\begin{aligned}
\label{e20}
\dot{\bar{V}}\leq&  -\rho_R \bar{V}_R - \rho_L \bar{V}_L \leq -\rho \bar{V}
\end{aligned}
\end{equation}
where $\rho = \min \hspace{0.1cm} [\hspace{0.1cm} \rho_R,\hspace{0.2cm}{\rho_L}\hspace{0.1cm}]=\min\limits_{i=R,L} \hspace{0.1cm} [\hspace{0.1cm} {g}_i \gamma_i,\hspace{0.2cm}{2\kappa_i}\hspace{0.1cm}]$. Thus, from Eq. \eqref{e20},
\begin{equation}
\small
\begin{aligned}
\label{e21}
\bar{V} (t) \leq&  \bar{V}(t_0) e^{-\rho(t-t_0)}
\end{aligned}
\end{equation}
From Eq. \eqref{e19}, we can easily obtain
\begin{equation}
\small
\begin{aligned}
\label{73}
\|\bm{e}\|^2 \leq&  2\bm{g}^{-1}_{min} \bar{V}(t_0) e^{-\rho(t-t_0)}
\end{aligned}
\end{equation}
where $\bm{g}_{min}$ is the minimum eigenvalue of the the matrix $\bm{g}$. Finally, we can prove \textit{Theorem 1}, as:
\begin{equation}
\small
\begin{aligned}
\label{73}
\|\bm{e}\| \leq&  \sqrt{2\bm{g}^{-1}_{min} \bar{V}(t_0)} e^{-\frac{\rho(t-t_0)}{2}}
\end{aligned}
\end{equation}
As $\bm{g}^{-1}_{min} \bar{V}(t_0)$ is always greater than or equal to zero, and based on \textit{Definition 1}, the proposed NN-RMFC ensures the global exponential stability of the studied SSHDR.

\section{A Guide to Tune Control Parameters}
\begin{figure}[h!] 
    \centering
    \scalebox{1.00}{\includegraphics[trim={0cm 0.0cm 0.0cm 0cm},clip,width=\columnwidth]{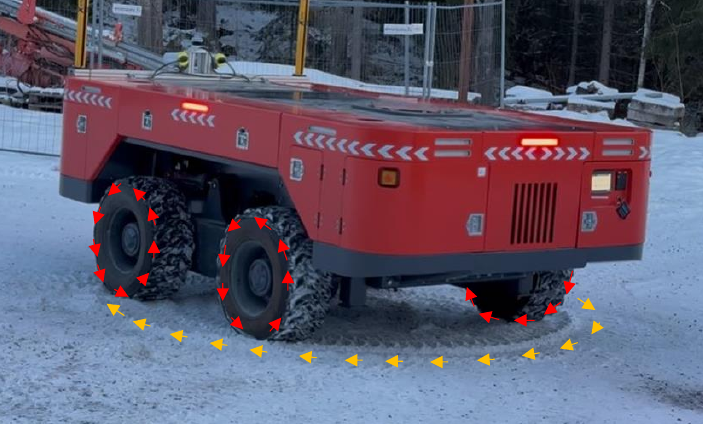}}
    \caption{Second round of tuning control design parameters; the video is available at: \texttt{https://youtu.be/JOh8lmQn2Rk}.}
    \label{fig45678g}
\end{figure}

The proposed NN-RMFC has the design parameters $\kappa_i$, $\epsilon_i$, $\sigma_i$, $\gamma_i$, $L_i$, $\varepsilon_{i,k}$, and $\alpha_{i,k}$ which require tuning. Based on our knowledge and experience, one of the best and safest ways to manually or automatically tune the design parameters of the proposed controller in SSHDRs in practice is as follows: 1) First round of tuning: this can be performed when there is no command motion ($V_{id} =0$) while the entire system is activated. In this state, approximate but imprecise control parameters can be determined by ensuring that the robot remains completely stationary.
2) Second round of tuning: the robot platform is commanded to perform smallest radius turns by moving one set of wheels forward and the other in reverse, allowing it to pivot within its own length ($V_R = - V_L$), as illustrated in Fig. \ref{fig45678g}.

The smaller the circle radius it can mechanically and structurally achieve, the closer we are to meeting the second-round tuning requirements and achieving suboptimal control design parameters. This step is recommended because any deviation from the small circular movement is easily observable by the operator, allowing for immediate refinement of tuning in case of an emergency.
3) Final tuning: after fulfilling the second-round requirements, the gains are adjusted to match the suboptimal values closely. However, for further accuracy, the robot should be commanded to follow different motion profiles and repeat the previous rounds to minimize tracking error by comparing the sensor measurements with the intended commands.

\section{Performance Validation and Results}
\subsection{Simulation Results in Slippery Terrains}
Building upon the work of \cite{petrovic2022analytic}, an external, independent method for self-assessment is employed in this study. Specifically, the Simscape Multibody${ }^{\mathrm{TM}}$ simulation environment is utilized to validate the proposed control scheme and to examine the impact of the introduced slip condition on the final solution. The simulation focuses on a four-wheel SSHDR with a differential drive system (two wheels on each side), where the selection of chassis configurations is flexible, as the governing equations are not constrained to any particular design or set of physical properties.
As shown in Table \ref{slip}, derived from \cite{yu2020dynamic, zhu2021effects, kristalniy2022friction, canudas1999model, zeng2020experimental}, during the execution, we considered randomly non-functional values of tires' slip ratios under different environmental conditions for Eqs. \eqref{e2} and \eqref{e1}. Although these are approximations, they are sufficient to demonstrate the control strategy's effectiveness.

\begin{figure}[h!] 
    \centering
    \scalebox{1.00}{\includegraphics[trim={0cm 0.0cm 0.0cm 0cm},clip,width=\columnwidth]{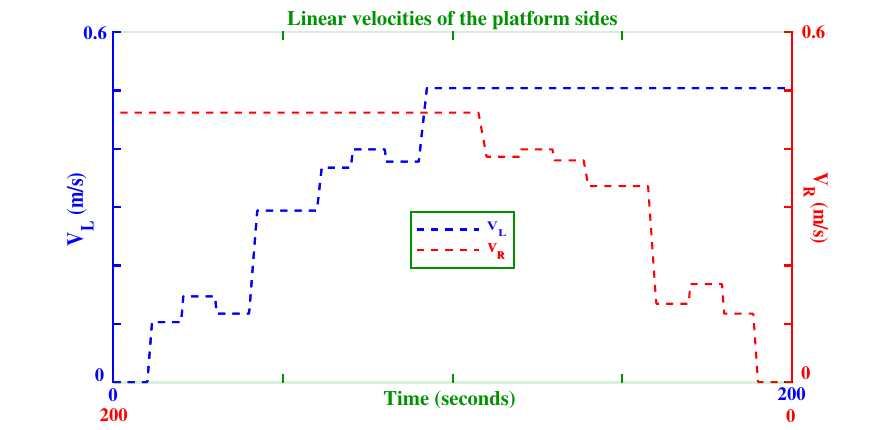}}
    \caption{Reference velocities for the robot platform.}
    \label{fig3asdas3}
\end{figure}

\begin{table}[h]
    \centering
    \caption{Considered Ranges Of Slip Ratio On The Platform For Various Surfaces}
    \label{slip}
    \small
\begin{tabular}{l c c}
\hline
\textbf{Surface type} & \textbf{Slip ratio of the left} & \textbf{Slip ratio of the right}\\
\hline
Dry asphalt & $0.05-0.40$ & $0.05-0.20$\\
Wet asphalt & $0.05-0.80$ & $0.05-0.50$ \\
Gravel & $0.05-0.5$ & $0.05-0.40$\\
Mud & $0.05-0.70$ & $0.05-0.50$ \\
Ice & $0.05-0.90$ & $0.05-0.75$\\
\hline
\end{tabular}
\end{table}
The reference velocities for each side were chosen differently to allow the robot to follow a curved path, as shown in Fig. \ref{fig3asdas3}. 

\begin{figure}[h!] 
    \centering
    \scalebox{1.00}{\includegraphics[trim={0cm 0.0cm 0.0cm 0cm},clip,width=\columnwidth]{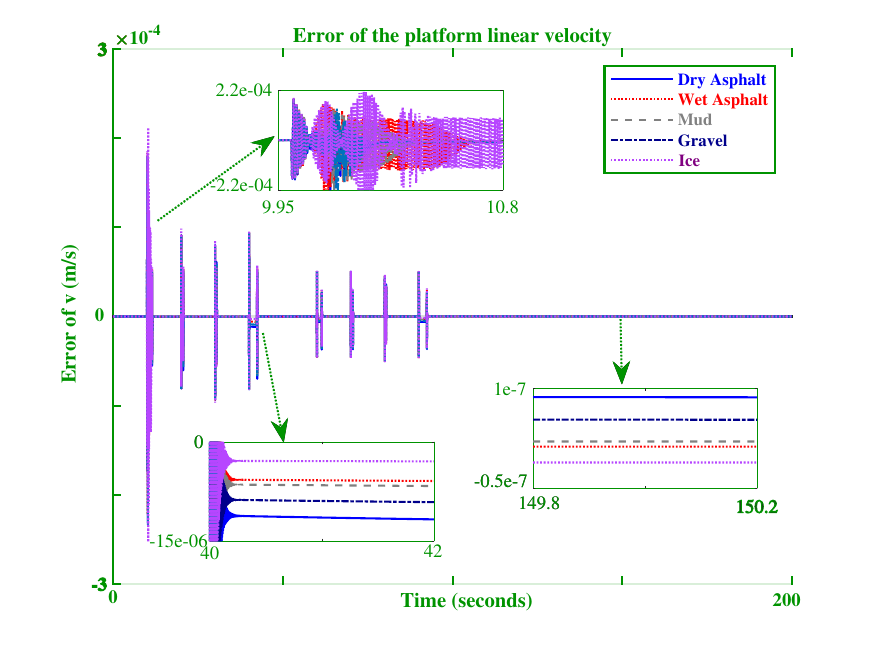}}
    \caption{{Velocity tracking error of the studied SSHDR in different terrains.}}
    \label{figcv33}
\end{figure}

The proposed control system including Eqs. \eqref{e5}--\eqref{e8} was applied to the simulated model, considering slip ratios provided in Table \ref{slip}, while the control design parameters were chosen as follows: $\kappa_i =1.2$, $\epsilon_i =0.04$, $\sigma_i = 11.5$, $\gamma_i = 1.6$, $L_i = 9$, $\varepsilon_{i,k} = [2 \hspace{0.1cm} {rand}-1, 2 \hspace{0.1cm} {rand}-1]^{\top}$, and $\alpha_{i,k}=0.13$, where $rand$ outputs a variable between $0$ and $1$.
The platform tracking errors for different operational surfaces are shown in Fig. \ref{figcv33}, validating the efficacy of the proposed control. This figure illustrates that the control-implemented robot exhibited a higher overshoot and longer settling time on high-slip-ratio surfaces while achieving a slightly lower steady-state error after the settling time. The zoomed-in section of the figure shows that dry asphalt had the fastest settling time, followed by gravel, mud, and wet asphalt, respectively, with the icy surface having the slowest settling time.

\begin{figure}[h!] 
    \centering
    \scalebox{1.00}{\includegraphics[trim={0cm 0.0cm 0.0cm 0cm},clip,width=\columnwidth]{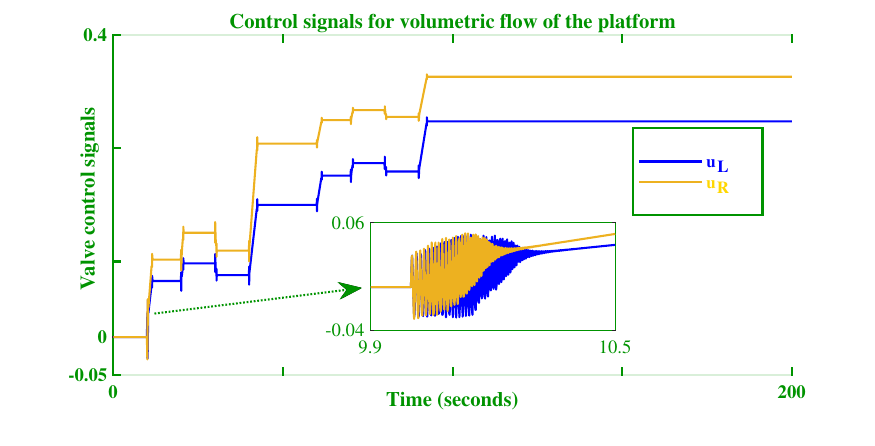}}
    \caption{{Control command signals.}}
    \label{fidasdasdg33}
\end{figure}

The control valve signals of the two serial hydraulic motors on each side are shown in Fig. \ref{fidasdasdg33}, illustrating that the control effort after 100 seconds remained relatively constant because the commanded velocity on both sides was constant.

\subsection{Experimental Results in Extremely Slippery and Snowy Terrains}
\begin{figure}[h!] 
    \centering
    \scalebox{1.00}{\includegraphics[trim={0cm 0.0cm 0.0cm 0cm},clip,width=\columnwidth]{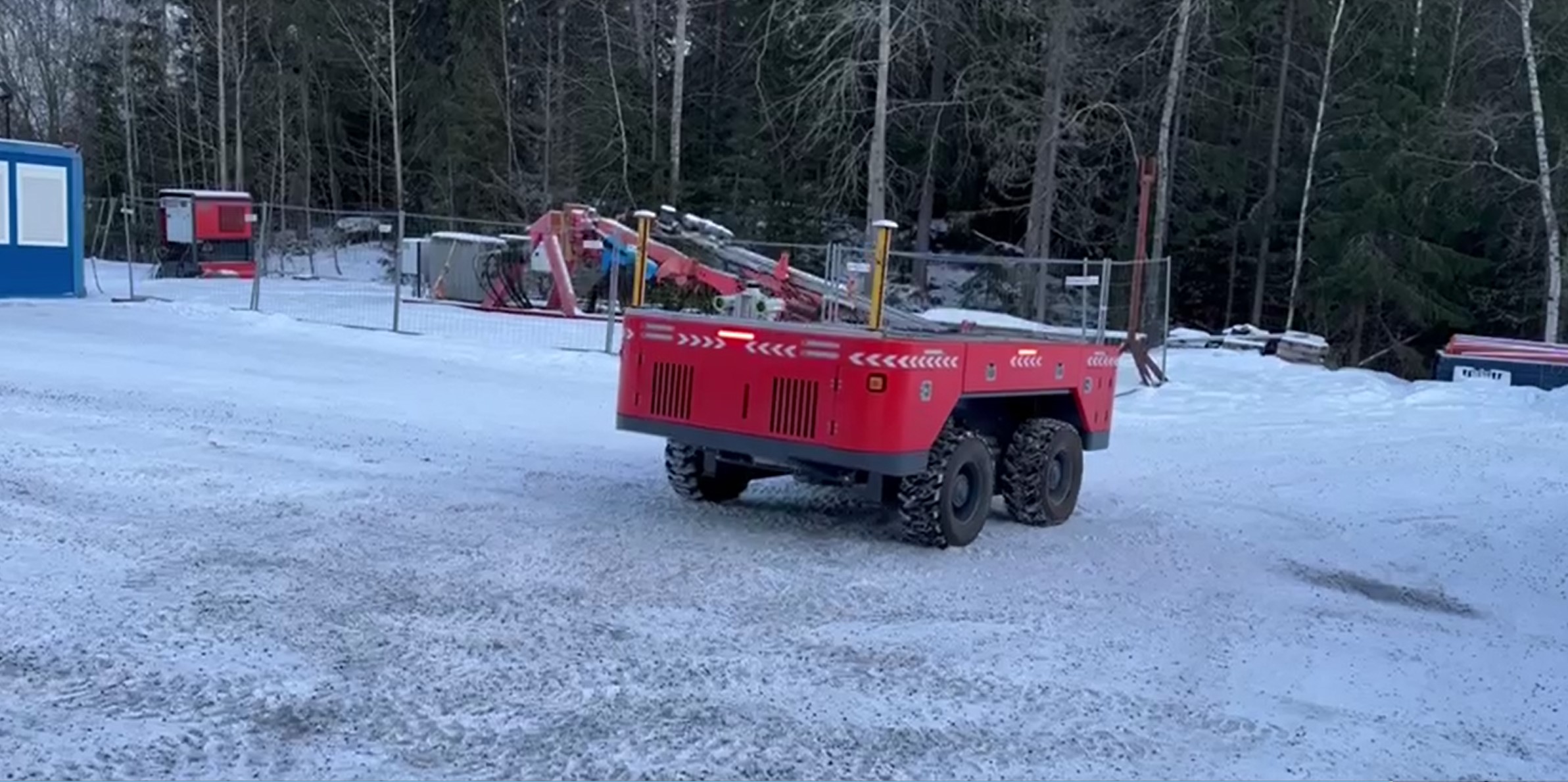}}
    \caption{Use-case SSHDR operating on the snowy terrain; for further information, the video is available at: \texttt{https://youtu.be/2q3NxWM5Soc}.}
    \label{fifdg45678g}
\end{figure}

In this section, the performance of the proposed control strategy is investigated when applied to a $4,836$ kg SSHDR. Each side of the robot, utilizing a differential drive mechanism, is independently actuated by a permanent magnet synchronous motor (PMSM) that drives a hydraulic pump, supplying fluid flow to a pair of hydraulic motors connected to the wheels via gearboxes. The electric system consists of a battery-powered inverter that regulates the speed and torque of each PMSM with precision. The hydraulic system transmits power from the pumps to the motors, with an oil tank maintaining fluid supply and system pressure. The mechanical system facilitates motion through skid-steering, where differential RPM control of each PMSM enables linear movement, turning, and pivoting. By adjusting the PMSM speeds, the controller ensures precise maneuverability, making the system highly adaptable for off-road and industrial applications, where hydraulic transmission provides high torque and efficiency, while electric control optimizes energy utilization. Each wheel is equipped with an odometry sensor, and the odometry outputs of the two wheels on each side are summed and divided by two to compute the measured velocity of that side at any given time step. The controller is designed to adjust the RPM of the PMSM sufficiently to provide precise maneuverability and track the reference motion commanded by a human-assisted operator through a joystick while ensuring the system remains highly adaptable and robust on snowy and slip-susceptible surfaces. It has a wheelbase of $1.85$ m, while we assume that the wheel diameter and gear ratio are unknown to the proposed controller. The manual tuning of the controller parameters was earlier shown in Fig. \ref{fig45678g}. The controller-implemented use case, SSHDR, was intended to operate on extremely slippery and snowy terrain, as shown in Fig. \ref{fifdg45678g}.

\begin{figure}[h!] 
    \centering
    \scalebox{0.95}{\includegraphics[trim={0cm 0.0cm 0.0cm 0cm},clip,width=\columnwidth]{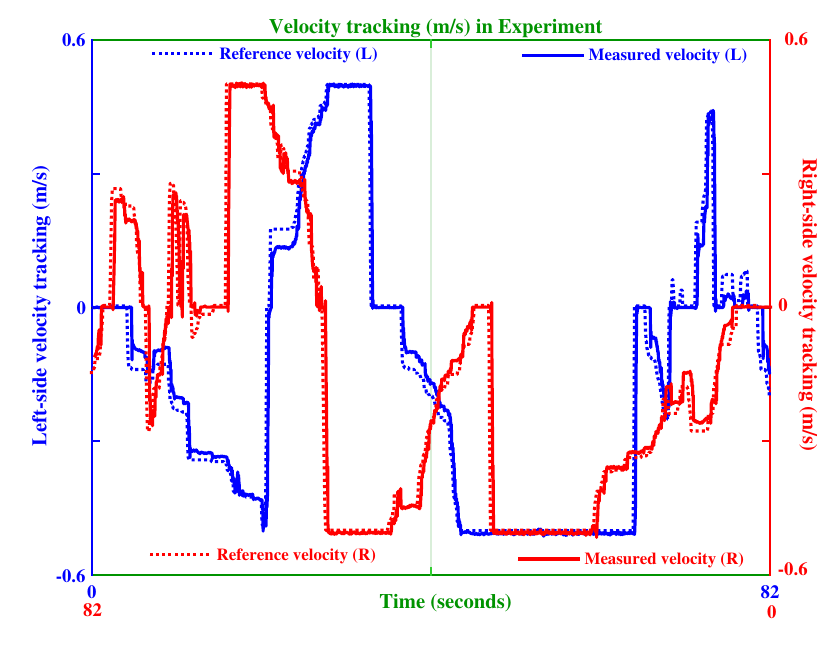}}
    \caption{{Velocity tracking error of the studied SSHDR under snowy terrain.}}
    \label{fig1csa33}
\end{figure}

The control design parameters were chosen as: $\kappa_i =1.9$, $\epsilon_i =0.08$, $\sigma_i = 17.1$, $\gamma_i = 3.6$, $L_i = 8$, $\varepsilon_{i,k} = [2 \hspace{0.1cm} {rand}-1, 2 \hspace{0.1cm} {rand}-1]^{\top}$, and $\alpha_{i,k}=0.15$, where $rand$ outputs a variable between $0$ and $1$.
Fig. \ref{fig1csa33} illustrates the tracking control
performance by depicting the actual measured velocity of each side, commanded by the proposed controller, along with the reference velocity generated by a human-assisted joystick. Note that a difference in the sign of the velocity indicates that the robot is pivoting, as the operator demands.

\begin{figure}[h!] 
    \centering
    \scalebox{1}{\includegraphics[trim={0cm 0.0cm 0.0cm 0cm},clip,width=\columnwidth]{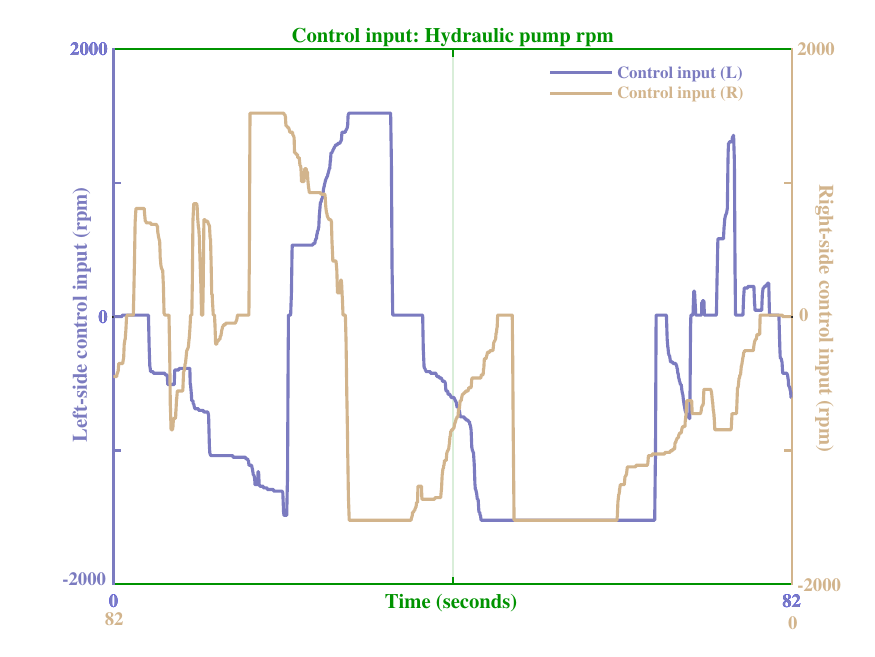}}
    \caption{{Control command signal, RPMs of PMSMs on two sides.}}
    \label{fig13wdxqq3}
\end{figure}

\begin{figure}[h!] 
    \centering
    \scalebox{1}{\includegraphics[trim={0cm 0.0cm 0.0cm 0cm},clip,width=\columnwidth]{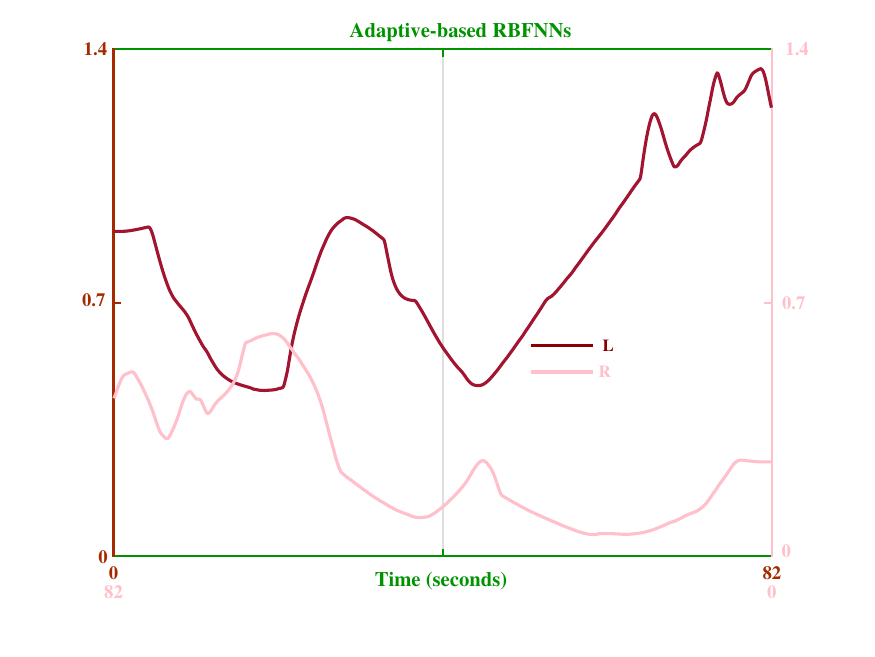}}
    \caption{{Updated values of the RBFNNs ($\|\Phi_i\|$).}}
    \label{fig13adadfecqw3}
\end{figure}

Fig. \ref{fig13wdxqq3} illustrates the RPM of PMSM during execution, which is commanded by the proposed controller to track the desired motion. Fig. \ref{fig13adadfecqw3} shows the updated values of RBFNNs on both sides. As observed in both the simulation and experimental results, the proposed NN-RMFC adheres to robust and anti-slip control principles for the studied SSHDR. To further investigate the controller's performance, four state-of-the-art controllers were applied to the studied SSHDRs.

The comparative results in Table II indicate that, overall, the proposed NN-RMFC demonstrates better performance for the robot's multi-stage actuation system. As reported in the table, the model-free adaptive predictive control (MFAPC) provided in \cite{zhu2024backstepping} was also applied to the studied SSHDR. While it also did not require any modeling information and demonstrated high robustness against snowy surface conditions, the performance of the NN-RMFC, particularly in terms of steady-state error, surpassed it. Additionally, MFAPC required the implementation of multiple algorithms, including adaptive control, model predictive control, and PID control, which made its implementation somewhat challenging.
On the other hand, the most common model-free control method, PID, was also applied. Although its implementation was straightforward for the studied SSHDR, it exhibited weak performance under the highly nonlinear conditions of the snowy surface.
In addition, the proposed controller in \cite{ge2023prescribed}, a fractional-order prescribed-time controller (FOPTC), was implemented in the studied SSHDR under the same conditions. The performance of the FOPTC-implemented SSHDR was competitive with the NN-RMFC, but it required a long settling time. Unfortunately, with predefinition of a better performance, FOPTC could theoretically extend indefinitely, which led to crashes in the computing system during the experiment.

\begin{table}[ht]
\small
\centering
\caption{Control Performance Metrics In The Experiment}
\begin{tabular}{|c|c|c|c|c|}
\hline
\textbf{Stability Criteria} & NN-RMFC & MFAPC & PID & FOPTC\\
\hline
\textbf{Settling time} & $1.2$ s & $1.6$ s & $2.8$ s & $3.3$ s\\
\textbf{Overshoot}  & $1.4 \%$ & $1.7\%$ &  $1.9\%$ & $1.2\%$\\
\textbf{Steady-state error} & $0.002$& $0.008$  & $0.035$& $0.004$ \\
\hline
\end{tabular}
\end{table}

\section{Conclusion}
To take a step toward reliable autonomous navigation and control in heavy-duty machinery undergoing robotization, this paper aimed to design an innovative NN-RMFC system to strongly stabilize the dynamics of a four-wheel SSHDR with a highly-nonlinear multi-stage actuation mechanism in the presence of a broad range of potential wheel slippage on different off-road surfaces. To guarantee global exponential stability of the SSHDR, the proposed controller was designed as follows: 1) the unknown modeling of wheel dynamics is approximated using radial basis function neural networks (RBFNNs); 2) a new adaptive law is proposed to compensate for slippage effects and tune the weights of the RBFNNs online during execution. Simulation and experimental results, for a $4,836$ kg SSHDR, verified the efficacy of the proposed control, operating on extremely slippery and snowy terrains.

\bibliographystyle{IEEEtran}
\bibliography{lcsys}

\end{document}